\documentclass[final]{cvpr}

\usepackage[accsupp]{axessibility}
\usepackage{times}
\usepackage{epsfig}
\usepackage{graphicx}
\usepackage{amsmath}
\usepackage{amssymb}
\usepackage{booktabs}
\usepackage{float}
\usepackage{url}
\usepackage{xcolor}
\usepackage{stfloats}
\usepackage{graphicx,float}
\usepackage{pifont}% http://ctan.org/pkg/pifont

\usepackage[pagebackref=true,breaklinks=true,colorlinks,bookmarks=false]{hyperref}

\def\cvprPaperID{} % *** Enter the CVPR Paper ID here
\def\confYear{CVPR 2021}

\begin{document}

%%%%%%%%% TITLE
\title{CLIP-Art: Contrastive Pre-training for Fine-Grained Art Classification}

\author{Marcos V. Conde, Kerem Turgutlu\\
{\tt\small drmarcosv@protonmail.com, keremturgutlu@gmail.com}
}

\maketitle

%%%%%%%%%%%%%%%%%%%%%%%%%%%%%%%%%%%%%%%%%%%%%%%%%%%%%%%%%%%%%%%%%%%%%%%%%%%%%%%

\thispagestyle{empty}
%\pagenumbering{gobble}

%%%%%%%%% ABSTRACT
\begin{abstract}
Existing computer vision research in artwork struggles with artwork's fine-grained attributes recognition and lack of curated annotated datasets due to their costly creation.
To the best of our knowledge, we are one of the first methods to use CLIP (Contrastive Language-Image Pre-Training) to train a neural network on a variety of artwork images and text descriptions pairs. CLIP is able to learn directly from free-form art descriptions, or, if available, curated fine-grained labels.
Model’s zero-shot capability allows predicting accurate natural language description for a given image, without directly optimizing for the task. 
Our approach aims to solve 2 challenges: instance retrieval and fine-grained artwork attribute recognition.
We use the iMet Dataset, which we consider the largest annotated artwork dataset. In this benchmark we achieved competitive results using only self-supervision. Our code is available at:
{\small{\url{https://github.com/KeremTurgutlu/clip_art}}}
\end{abstract}

%%%%%%%%% TEASER FIGURE

%\thispagestyle{empty}

\section{Introduction}
\label{introduction}

How to tell in which culture a sculpture was made? There are hundreds of possibilities: Greek, Roman, Arabic, and more.
Fine-Grained Visual Classification (FGVC) aims to classify
the sub-categories under coarse-grained large categories,
such as the author of a painting, material of a sculpture,
country of origin of an instrument. FGVC is challenging
because objects that belong to different categories might
have similar characteristics, but differences between subcategories
might be remarkable (small interclass variations
and large intra-class variations). Because of these reasons,
it is hard to obtain accurate classification results using
classical Convolutional Neural Networks \cite{imagenet, resnet, inception, vgg}.
Recent work \cite{ Lam_2017_CVPR_fgvc, Chen_2019_CVPR_fgvc, zhang2020multibranch, conde2021exploring}  shows the key step of FGVC is identifying and extracting more informative regions and features in an image. 
However, labeling fine-grained categories is an expensive and time-consuming process which often requires expertise in a specialized domain, thus, FGVC datasets \cite{data_KhoslaYaoJayadevaprakashFeiFei_FGVC2011, data_WahCUB_200_2011, data_maji2013finegrained} have limited training data.
\begin{figure}[ht!]
    \begin{center}
    \includegraphics[width=\linewidth]{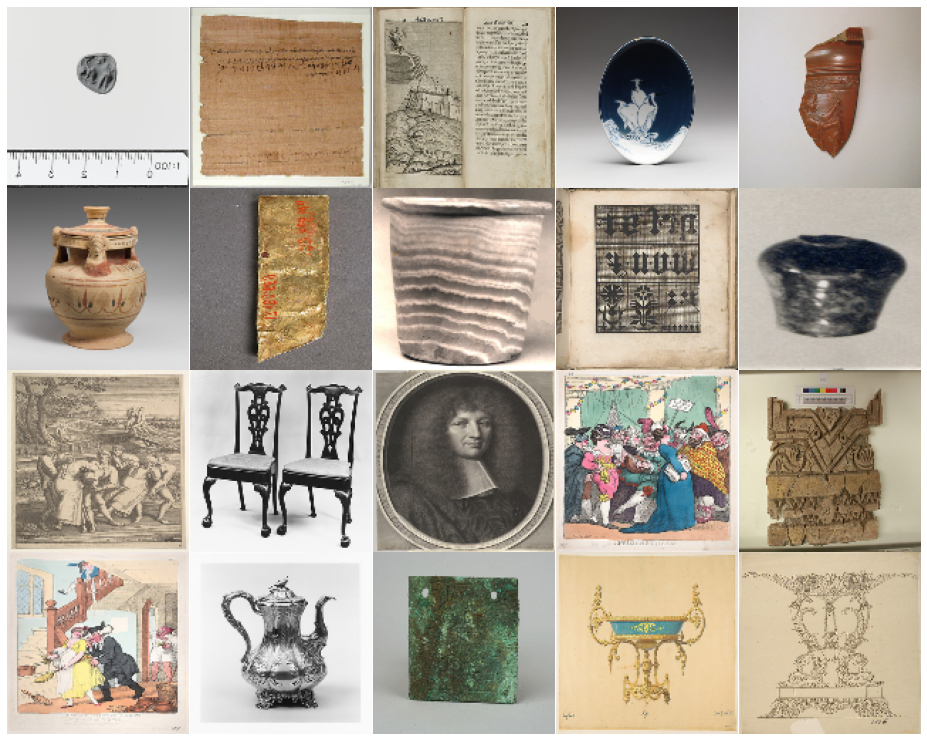}
    \end{center}
    \caption{The artworks in iMet \cite{zhang2019imet} include paintings, instruments, prints, clothing, sculpture, furniture, metalwork, etc. }
    \label{fig:teaser}
\end{figure}
For this reason, research focuses on weakly-supervised learning using noisy labels, unsupervised and self-supervised learning schemes to recognize informative regions in the images~\cite{hu2019better_fgvc, yang2018learning_fgvc, fgvc_cvpr2020_details, fgvc-text}.

\textbf{Our main contributions are:}

\begin{itemize}

    \item Explore Contrastive Pre-training \cite{radford2021learning} framework for
    fine-grained visual-textual representation learning \cite{fgvc-text} by using natural language free-form descriptions of artwork and images.

    \item A multimodal representation learning for classification and image-text retrieval.
        
    \item Our task-agnostic model performs zero-shot fine-grained classification, and achieves better results than few-shot supervised SOTA models \cite{resnet, effnet}.

\end{itemize}

\subsection{Dataset and Benchmark}
\label{dataset}

\begin{figure}[ht!]
    \begin{center}
    \includegraphics[width=0.85\linewidth]{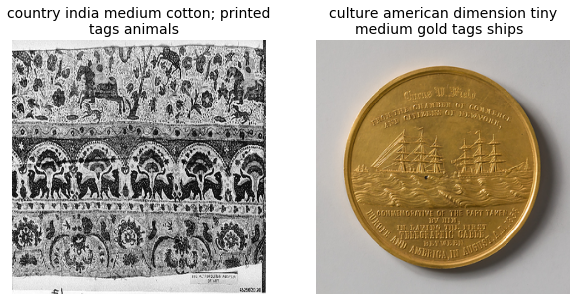}
    \end{center}
    \caption{Image and noisy fine-grained categories.}
    \label{fig:teaser_attr}
\end{figure}

The iMet Collection Dataset \cite{zhang2019imet} from The Metropolitan Museum of Art in New York (The Met), presents the largest fine-grained artwork collection. Some samples are shown in Figure \ref{fig:teaser}.
Each image is labeled with its associated artistic attributes. 
The attributes can relate to what one ``sees" in the work or what one infers as the object's ``utility".

Figure \ref{fig:teaser_attr} shows images and their attributes description. \\
These are grouped into 5 parent classes: country, culture, dimension, medium, tags. 
In total, there are 3471 unique attributes.
Research-grade Museum experts curated and verified attribute labels to ensure high quality. 
However, each object is annotated by a single annotator without a verification step, and sometimes they added free-form text descriptions. 
For this reason, the authors recommend considering attributes as \textbf{noisy} labels.
iMet hosts a yearly competition since 2019~\cite{imet-bench}, providing a public benchmark based on more than 40.000 unknown test images.

%\thispagestyle{empty}
%-------------------------------------------------------------------------
\section{Approach}
\label{approach}

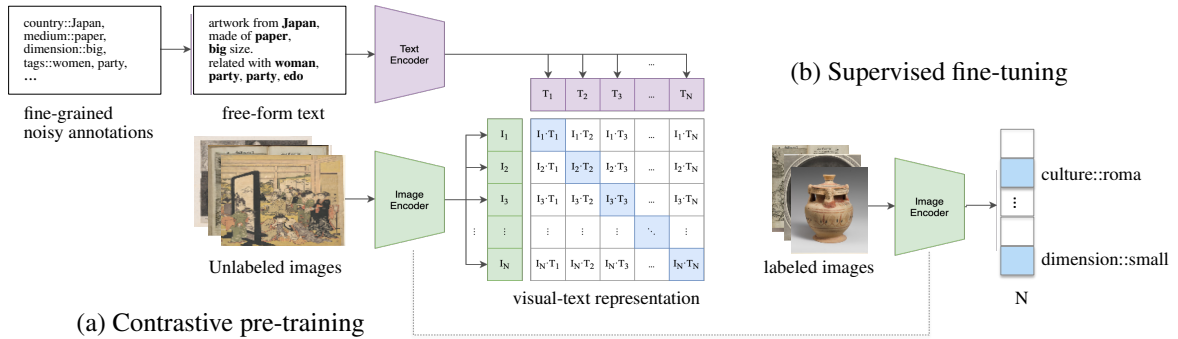
\begin{figure*}[hb!]
    \begin{center}
    \includegraphics[width=0.8\linewidth]{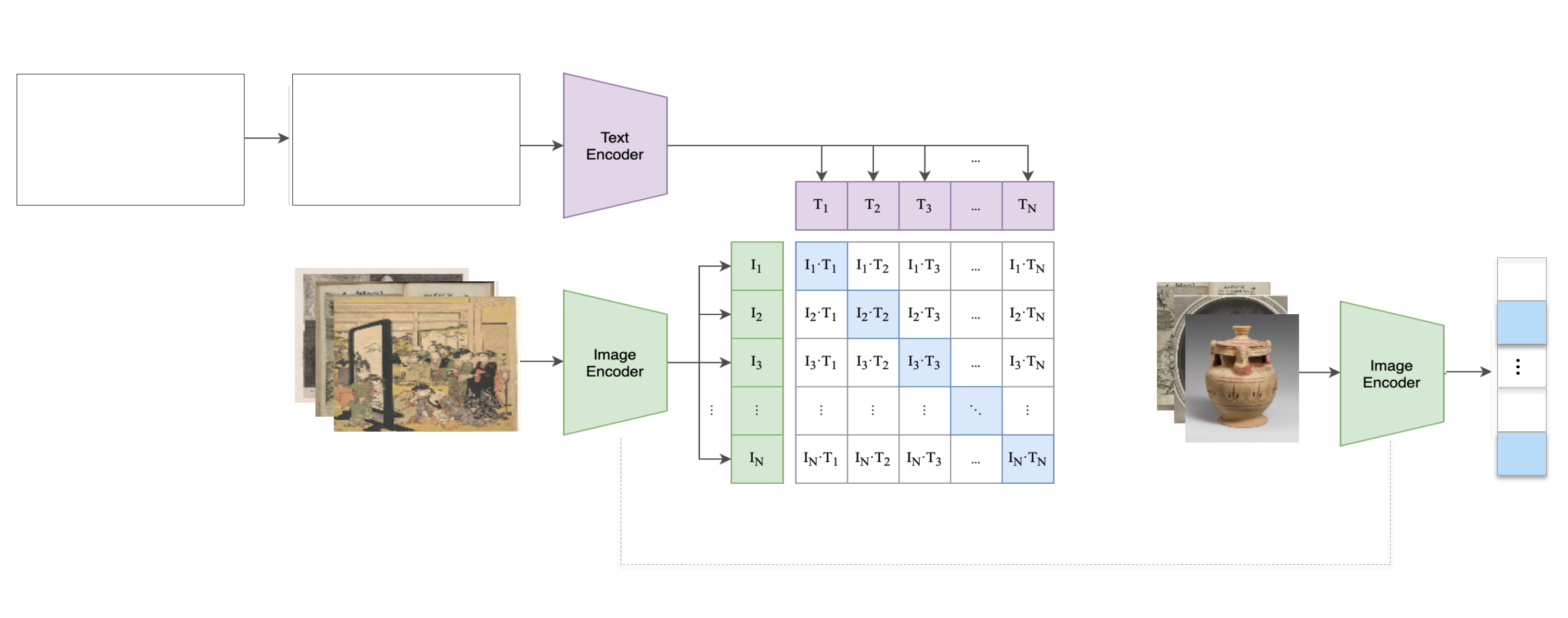}
    \put(-320,130){ \tiny artwork from \textbf{Japan}, }
    \put(-320,125){ \tiny made of \textbf{paper}, }
    \put(-320,120){ \tiny \textbf{big} size. }
    \put(-320,115){ \tiny related with \textbf{woman},  }
    \put(-320,110){ \tiny \textbf{party}, \textbf{party}, \textbf{edo} }
    \put(-390,130){ \tiny country::Japan, }
    \put(-390,125){ \tiny medium::paper, }
    \put(-390,120){ \tiny dimension::big, }
    \put(-390,115){ \tiny tags::women, party, }
    \put(-390,110){ \scriptsize ... }
    \put(-390,95){ \scriptsize fine-grained}
    \put(-390,88){ \scriptsize noisy annotations }
    \put(-315,95){ \scriptsize free-form text}
    \put(-320,37){ \scriptsize Unlabeled images }
    \put(-205,25){ \scriptsize visual-text representation }
    \put(-370,15){ \text (a) Contrastive pre-training }
    \put(-110,37){ \scriptsize labeled images }
    \put(-15,25){ \scriptsize N }
    \put(-5,72){ \scriptsize culture::roma }
    \put(-5,40){ \scriptsize dimension::small}
    \put(-100,110){ \text (b) Supervised fine-tuning }
    \end{center}
    \caption{Summary of our approach based on CLIP from OpenAI \cite{radford2021learning}.
    We show (a) Contrastive pre-training using unlabeled images (or noisy annotated). 
    We process noisy or scrapped annotations into natural language free-form descriptions as explained in Section \ref{approach}.
    Using a task-agnostic image encoder and text encoder, we learn a visual-textual representation, discovering discriminative visual-textual pairwise information \cite{fgvc-text}.
    Further supervised fine-tuning (b) can be done using small labeled datasets.}
    \label{fig:main}
\end{figure*}

Our approach consists of multiple stages which can be seen in Figure \ref{fig:main}; free-form text generation, contrastive pre-training and finally fine-tuning on the downstream fine-grained art recognition task. 
First, we convert noisy fine-grained categorical annotations into natural language text for a given image. We achieve this by using natural language templates and using different permutations. This process is similar to data augmentation but for text descriptions. 

We use different combinations of attribute values when there might be an image with multiple attribute values for a given category, such as multiple tags which can describe different things in an art object. 
At the end of this data generation process, we end up having more than 15 text descriptions per image in the iMet dataset~\cite{zhang2019imet}. 

Second, we fine-tune ViT-B/32 CLIP~\cite{vit,radford2021learning} model which is open-sourced by OpenAI. This model uses 2 transformer encoders for jointly embedding the text and image pairs; a ViT-B/32 for image encoding and another 12-layer transformer for text encoding. Similar to the original CLIP model \cite{radford2021learning} we minimize InfoNCE loss \cite{oord2019representation} during contrastive pre-training. 
In a given batch, each image-text pair or text-image pair forms a positive sample and every other image or text is considered negative. Having this symmetry we calculate pairwise cosine similarity between $\textit{L}_{2}\text{-normalized}$ image-text embeddings and calculate cross-entropy loss with a learnable temperature parameter.
In our synthetic dataset, a given image has multiple text descriptions, for that reason we randomly sample one text with equal probability during training. This can be viewed as data augmentation. Additionally, we apply dropout to attribute values if there are multiple values for a given category to further diversify these augmentations. For the remainder of this paper, we refer to OpenAI Vit-B/32 model as $\text{CLIP}_{base}$ and our fine-tuned version as $\text{CLIP}_{art}$. %
Finally, we use the domain adapted $\text{CLIP}_{art}$ for further fine-tuning on the downstream fine-grained art recognition task.

\subsection{Contrastive Pre-training}

\begin{figure*}[hb!]
    \begin{center}
    \includegraphics[width=0.85\linewidth]{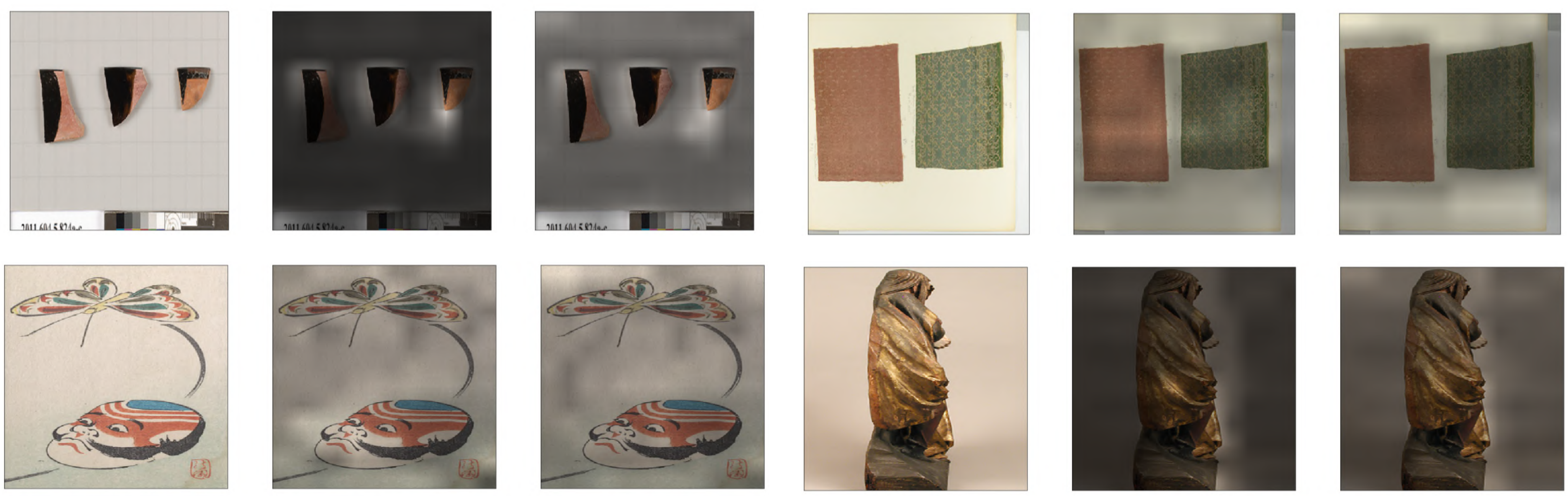}
    \end{center}
    \caption{Attention map samples. For each sample we show (left) image, (middle) attention map from CLIP-Art and (right) from CLIP-Base~\cite{radford2021learning}. Our contrastive learning of visual-text features helps to discriminate better the most discriminative regions in the image.}
    \label{fig:results-attention}
\end{figure*}

In our experiments, contrastive pre-training shows the following advantages: it can leverage free-form text to learn more generalized and robust visual features even in the presence of noise, it allows faster and better convergence for the downstream task at hand, it can be used for retrieval with any natural language query at inference time beyond the closed set of predefined labels and different loss functions from any state-of-the-art self-supervised learning method can be used \cite{chen2020simple, zbontar2021barlow} during training.
We fine-tuned models using Ranger optimizer, a combination of Lookehead and RAdam \cite{liu2020variance, zhang2019lookahead}. 
No image data augmentation is used besides random resized cropping and horizontal flip. All contrastive pre-training models are trained for 20 epochs. To test our hypothesis that free-form text descriptions help to learn good fine-grained representations, we fine-tuned $\text{CLIP}_{art}$ model with 2 different versions of text pairs; one which is generated using all 5 categories and another which does not include \textbf{tags} category. Later retrieval performance for these 2 versions are reported in Table \ref{retieval-results}.
We show examples of the corresponding attention maps in Figure~\ref{fig:results-attention}.

\subsection{Fine-tuning}

For all downstream fine-tuning experiments same setup;
image size, data augmentation, MLP layers, and learning
rate schedulers are used for a fair comparison. We treated the
downstream fine-grained art attribute recognition as a multilabel
classification task where each attribute is assumed
to be independent and an image can be assigned multiple
attributes as can be seen from Figure \ref{fig:main}. For the first 5 epochs encoder weights are frozen and for the remaining 15 epochs all model weights are updated.

%-------------------------------------------------------------------------

\section{Experiments}

In this section, we describe our experimental setup and results at fine-grained classification and artwork retrieval.
All CLIP-based models have as backbone a Visual Transformer (ViT) \cite{vit}. We conducted experiments for assessing the zero-shot, few-shot and fully supervised performance of variety of models including $\text{CLIP}_{base}$ and $\text{CLIP}_{art}$.

\subsection{Zero-Shot Experiments}
\label{zero-shot}

Using visual encoder, ViT-B/32, of $\text{CLIP}_{base}$  and $\text{CLIP}_{art}$ models we extracted image representations of 512-dimension for the full iMet 2020 training set, which consists of a total of 142,119 images. Later, we predicted on a 20K hold-out set using a query image and assigning the labels from the nearest neighbor in the training set. 

\thispagestyle{empty}
\subsection{Multimodal Retrieval Experiments}
\label{multi-retrieval}

In order to test our hypothesis that rich free-form text helps with learning better representations we train 2 versions of $\text{CLIP}_{art}$ using 2 different datasets with different text descriptions. 
We evaluate different versions of $\text{CLIP}_{art}$ with several retrieval metrics after encoding all the 20k validation images and their corresponding text descriptions. Once embedded we calculate normalized pairwise cosine similarity between all the image and text embeddings. Using this similarity matrix we report results in Table \ref{retieval-results} on retrieval percentage at 5, retrieval percentage at 20, mean retrieval rank, and median retrieval rank. 
%that removing such descriptive text hurt the retrieval performance significantly.

\subsection{Few-Shot and Fully Supervised}

In few-shot experiments, we trained models with a fraction of data to compare against the zero-shot performance. In fully supervised experiments we used full training data and compared a variety of models including ViT-B/32 from  $\text{CLIP}_{art}$, ViT-B/32 from $\text{CLIP}_{base}$, ViT-B/32 pre-trained on ImageNet~\cite{imagenet_data}. As well as a variety of ResNets~\cite{resnet} and EfficientNets~\cite{effnet} for benchmarking.

\subsection{Results}

\begin{figure*}[ht!]
    \begin{center}
    \includegraphics[width=0.9\linewidth]{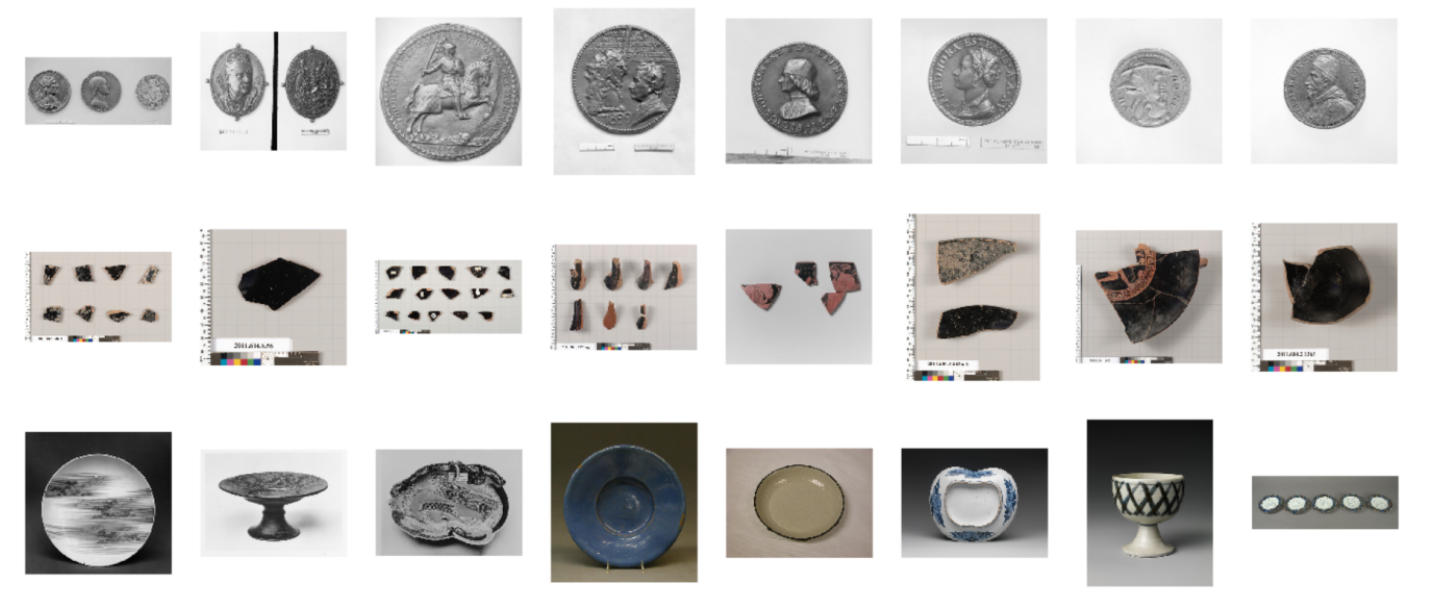}
    \put(-460,150){(a)}
    \put(-460,90){(b)}
    \put(-460,30){(c)}
    \end{center}
    \caption{Multimodal Retrieval results. Each row is the result of the following text queries:
    (a) \textit{``italian, rome artwork with portraits, profiles, men, popes made from bronze",}
    (b) \textit{``art from greek, attic made of terracotta"} and
    (c) \textit{``small japan artwork made from matsugatani type with dishes"}.
    Note that these images are completely unknown for our model.
    We obtain these results as explained at Section \ref{multi-retrieval}.
    %Quantitative results at Table \ref{retieval-results}
    }
    \label{fig:results2}
\end{figure*}

\begin{table}[!ht]
    \begin{center}
    \begin{tabular}{l c c c}
    \toprule
    Method & Backbone & Data (\%) & F2 score \\
    \midrule
    $\text{CLIP}_{base}$ \cite{radford2021learning} & ViT \cite{vit} & 0 & 0.5161 \\
    $\text{CLIP}_{art}$  (ours) & ViT \cite{vit} & 0 & \textbf{0.5507} \\
    ResNet \cite{resnet} & ResNet-50 & 10 & 0.5210 \\
    ResNet \cite{resnet} & ResNet-50 & 20 & 0.541 \\
    EfficientNet \cite{effnet} & EffNet-B0 & 10 & 0.511 \\
    EfficientNet \cite{effnet} & EffNet-B0 & 20 & 0.550 \\
    $\text{CLIP}_{art}$ (ours) & ViT \cite{vit} & 100 & 0.60 \\
    ResNet \cite{resnet} & ResNet-50 & 100 & \textbf{0.615} \\
    $\text{ViT}_{imet}$ \cite{vit} & ViT \cite{vit} & 100 & {0.58} \\
    \bottomrule
    \end{tabular}
    \end{center}
    \caption{Ablation study of the proposed methods. Data 0\% corresponds to zero-shot experiments, 10-20\% corresponds to few-shot and 100\% corresponds to completely supervised.}
    \label{classification}
\end{table}

\begin{table}[!ht]
    \begin{center}
        \resizebox{\columnwidth}{!}{\begin{tabular}{l c c c c}
            \toprule
            Dataset & ret\@5 & ret\@20 & mean ret & median ret \\
            \midrule
            $\text{All Categories}$ & \textbf{0.3052} & \textbf{0.5467} & 175.84 & 16  \\
            $\text{All (no ``Tags")}$ & 0.1658 & 0.3578 & 353.99 & 48  \\
            \bottomrule
        \end{tabular}}
    \end{center}
    \caption{Retrieval Results. We tested that removing a highly descriptive category such as  \textbf{tags} hurts retrieval performance and supports that representations when learned conditionally on descriptive text help with fine-grained retrieval. See Section~\ref{multi-retrieval}.}
    %We evaluate on 20K validation images for retrieval.
    \label{retieval-results}
\end{table}

We report results for zero-shot, few-shot and fully supervised training using the iMet dataset \cite{zhang2019imet}. We calculate F2-score metric to provide some robustness against noisy labels, favoring recall over precision. We use the validation consisting of 20K images.\\

In \textbf{zero-shot} benchmarks we used the KNN approach explained in Section~\ref{zero-shot}. For \textbf{few-shot} benchmarks, only 10\% and 20\% random subset of the training data is used for training classification SOTA CNNs~\cite{effnet,resnet}.
Table \ref{classification} shows that {$\text{CLIP}_{art}$}, a task-agnostic model, without any supervision outperforms ResNet-50 \cite{resnet} and EfficientNet-B0 \cite{effnet}, both SOTA classification models trained with a fraction of the complete dataset and optimized for the task.
Moreover, at a complicated fine-grained categorization task.
We show that a simple fine-tuned vision transformer can achieve results as full-supervised CNNs.
Furthermore, our {$\text{CLIP}_{art}$} ViT achieves better results than ViT pre-trained on ImageNet~\cite{imagenet_data}, and in half number of epochs. 
Note that the idea of this work is to explore multi-modality and image-text representations, for this reason, we do not use complex models, ensembles, aggressive augmentations, etc, as many solutions for this benchmark propose. More information at the appendix.
We show our multi-modality capability at Table \ref{retieval-results}. Our model is able to get the correct complete text pair for a given query image within the first 20 ranked predictions out of 20,000 candidates for the 54\% of the time. 
Table \ref{retieval-results} also shows that removing such descriptive text hurt the retrieval performance significantly. Qualitative results using images as queries can be found at the appendix.

%%%%%%%%%%%%%%%%%%%%%%%%%%%%%%%%%%%%%%%%%%%%%%%%%%%%%%5

%\thispagestyle{empty}
\section{Conclusion}

To solve art-related computer vision main challenges, retrieval and fine-grained attribute recognition, we present an approach based on Contrastive Language-Image Pre-Training (CLIP) using a wide variety of artwork images and natural language supervision. 
By its design, the network can be instructed in natural language to perform fine grained artwork retrieval and recognition in a zero-shot manner without directly optimizing for the iMet data. We also proposed a way for constructing natural language text from the available closed set of attribute labels by augmenting them. We hope this work represents a breakthrough in artwork applications, and helps related generative models.

%Future work can focus on building an artwork dataset consisting of more than 1 million image-text pairs scrapped from iMet's database, which, together with this work, will represent a breakthrough in artwork classification and retrieval.

%%%%%%%%%%%%%%%

%%%% APPENDIX

\section*{A1. Training convergence}

We compare ViT-B/32 \cite{vit} fine-tuning for fine-grained art classification using the weights from three different pre-training strategies:

\begin{enumerate}
    \item base contrastive pretraining CLIP \cite{radford2021learning} on 400 million images, open-sourced by OpenAI.
    \item our $\text{CLIP}_{Art}$ contrastive pretraining using artwork images and their natural language descriptions,
    \item ImageNet pretraining \cite{imagenet, imagenet_data}.
\end{enumerate}

\begin{figure}[htp!]
    \begin{center}
    \includegraphics[width=\linewidth]{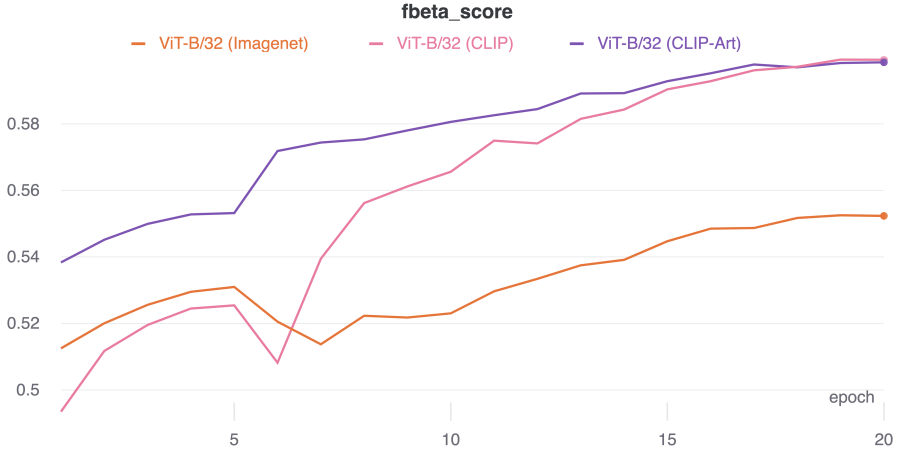}
    \end{center}
    \caption{
    Convergence plot for the first stage of training. Our $\text{CLIP}_{Art}$ improves convergence and performance (F2-score). For fair comparisons, we train to convergence using the same training setup (loss, optimizer, etc.) and images in all experiments.
    }
    \label{fig:training}
\end{figure}

%%%%%%%%% BODY TEXT

\section*{A2. Large Scale Art Dataset}
\label{semi}

\paragraph{Supervised CNNs}
We train more complex models for fine-grained art classification using 384 image size and exhaustive augmentations (random crops, horizontal and vertical flips, mixup). These models represent our set of teachers, their results on iMet \cite{zhang2019imet} are shown at Table \ref{tab:results2020lb}. Each teacher model trained with labeled data will infer pseudo-labels on unlabeled artwork data, which can be scrapped from the internet. This serves as dataset for self-training / distillation of task predictions using smaller versions of these models as noisy students~\cite{xie2020selftraining, NEURIPS2020_fcbc95cc}.

\begin{table}[ht!]
    \begin{center}
        \begin{tabular}{l c}
            \toprule
            Network & F2-score \\
            \midrule
            SEResNext-50 \cite{hu2019squeezeandexcitation}  & 0.701  \\
            EfficientNet-B7 \cite{effnet} & \textbf{0.712}  \\
            ViT-L-16 \cite{vit} & 0.707  \\
            \bottomrule
        \end{tabular}
    \end{center}
    \caption{Supervised SOTA CNNs trained on iMet dataset~\cite{zhang2019imet} and evaluated on 2020 Benchmark~\cite{imet-bench}.}
    \label{tab:results2020lb}
\end{table}

\begin{figure}[ht!]
    \begin{center}
    \includegraphics[width=\linewidth]{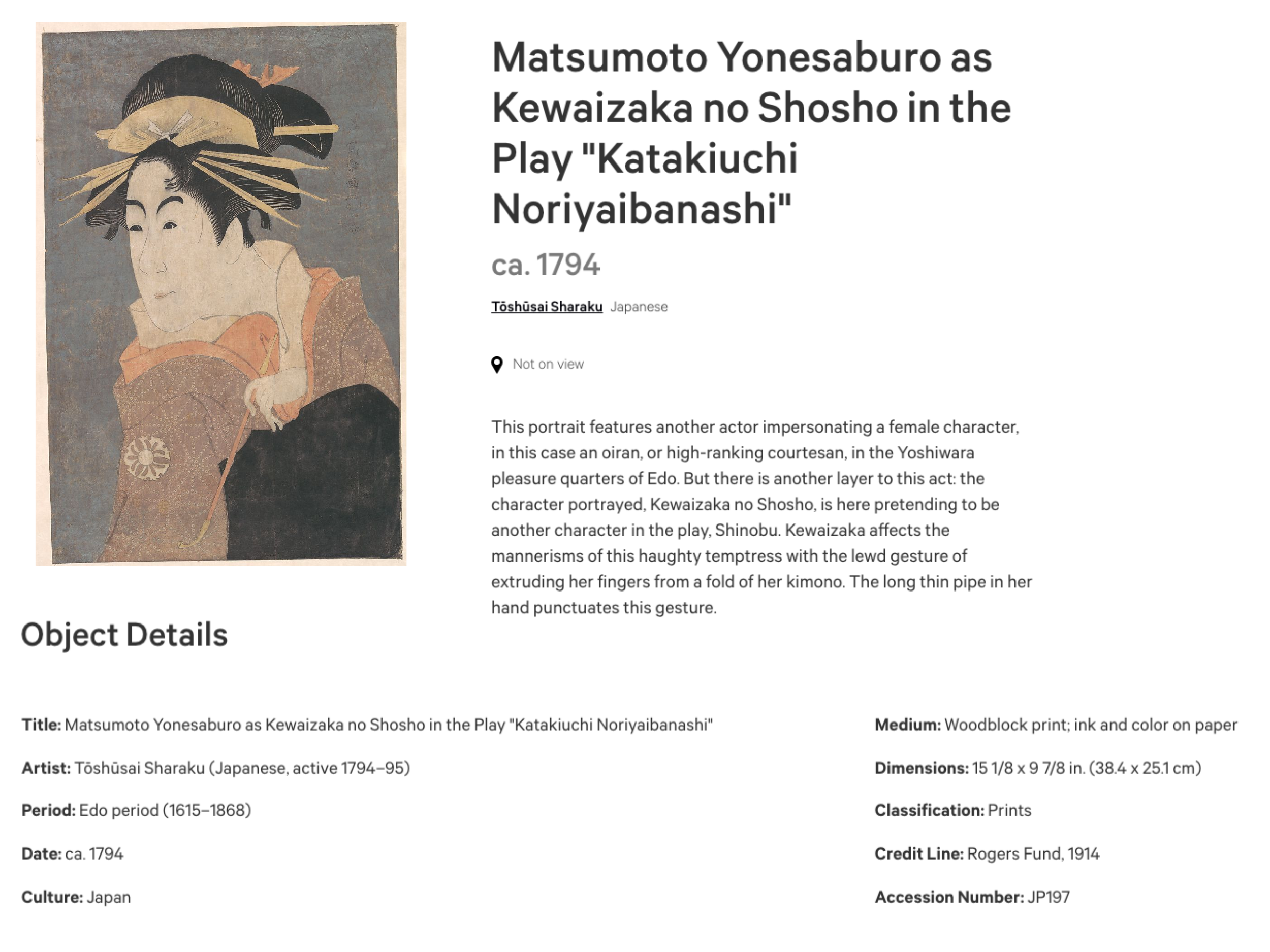}
    \end{center}
    \caption{Scrapped image and its natural language description. Source: \url{https://www.metmuseum.org/art/collection/search/36674}
    }
    \label{fig:scrap_sample}
\end{figure}

Scrapped images as Figure \ref{fig:scrap_sample} include an extensive free-form description from an expert, these are involuntary transferences from human visual attention to textual attention, which implies that textual attention can help to discriminate significant parts or features for categorization \cite{fgvc-text}.

\begin{figure*}[ht!]
    \begin{center}
    \includegraphics[width=\linewidth]{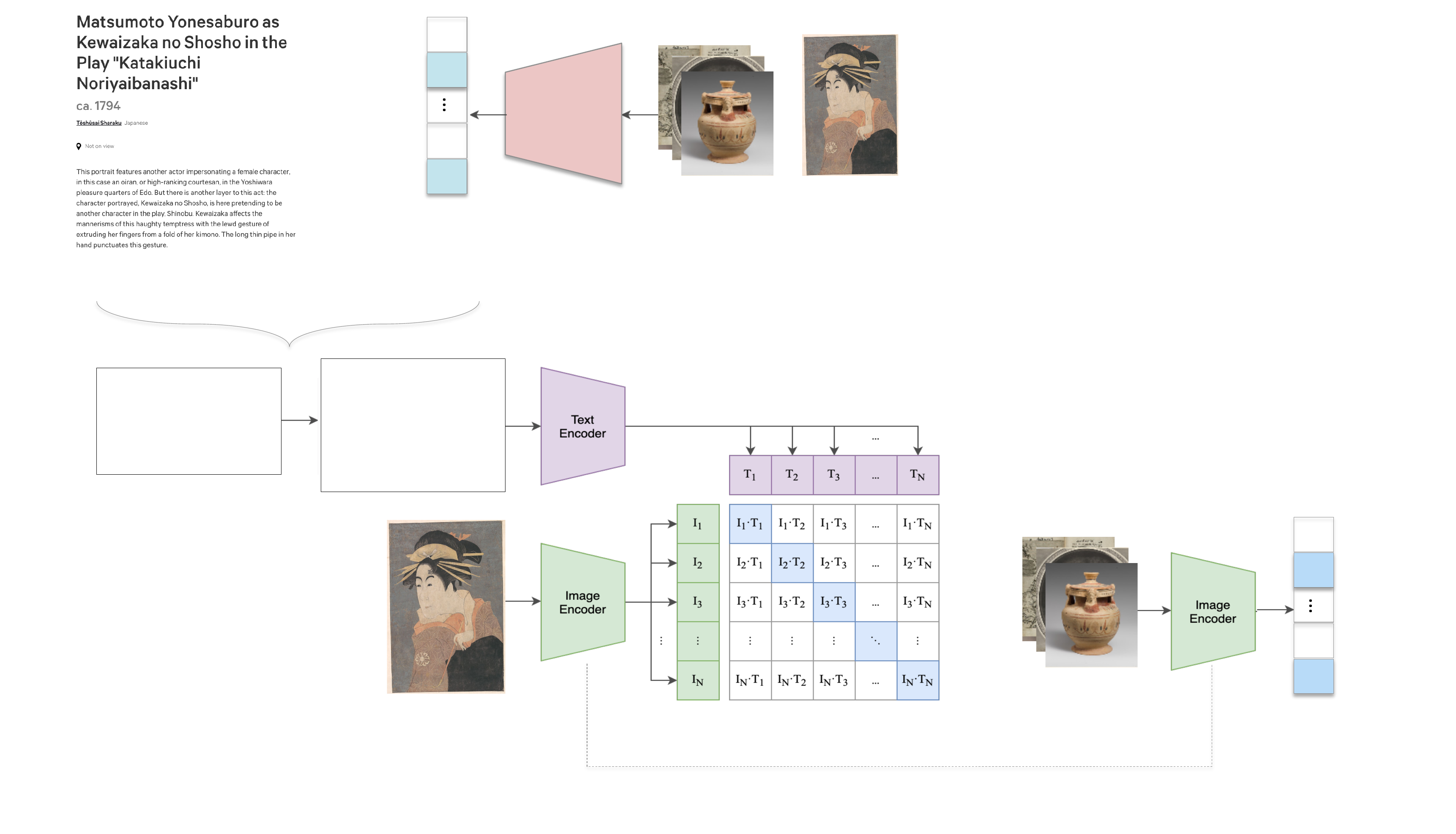}
    \put(-385,150){ \scriptsize artwork from \textbf{Japan}, }
    \put(-385,143){ \scriptsize made of \textbf{paper}, }
    \put(-385,136){ \scriptsize \textbf{big} size. }
    \put(-385,129){ \scriptsize related with \textbf{woman},  }
    \put(-385,122){ \scriptsize \textbf{party}, \textbf{party}, \textbf{edo} }
    \put(-463,150){ \scriptsize country::Japan, }
    \put(-463,143){ \scriptsize medium::paper, }
    \put(-463,136){ \scriptsize dimension::big, }
    \put(-463,129){ \scriptsize tags::women, party, }
    \put(-463,125){ \scriptsize ... }
    \put(-470,186){ \scriptsize natural language description + \scriptsize fine-grained pseudo-labels}
    \put(-350,210){ \scriptsize N }
    \put(-320,210){ Teacher }
    \put(-318,242){ CNN }
    \put(-272,210){ \scriptsize labeled images }
    \put(-225,210){ \scriptsize unlabeled images }
    \put(-370,37){ \scriptsize Unlabeled images }
    \put(-250,35){ \scriptsize visual-text representation }
    \put(-420,15){ \text (a) Contrastive pre-training }
    \put(-150,48){ \scriptsize labeled images }
    \put(-53,35){ \scriptsize N }
    \put(-40,86){ \scriptsize culture::roma }
    \put(-40,50){ \scriptsize dimension::small}
    \put(-150,130){ \text (b) Supervised fine-tuning }
    \end{center}
    \caption{Summary of our semi-supervised approach based on CLIP from OpenAI \cite{radford2021learning}.
    We show our teacher networks trained on iMet \cite{zhang2019imet} labeled data as explained in Section \ref{semi} and Table \ref{tab:results2020lb}. Scrapped text and pseudo-labels inferred from unlabeled images are processed into free-form descriptions.
    We also show (a) Contrastive pre-training using unlabeled images and their noisy generated descriptions. 
    Using a task-agnostic image encoder and text encoder, we learn a visual-textual representation, discovering discriminative visual-textual pairwise information \cite{fgvc-text}.
    Further supervised fine-tuning (b) can be done using labeled images.
    }
    \label{fig:diagram}
\end{figure*}

Using the teacher's predicted pseudo-labels and scrapped free-form descriptions from experts, we conform (image, text) pairs for learned visual attention from natural language supervision using CLIP \cite{radford2021learning}.
In this way, we aim to build an artwork dataset consisting of more than 1 million (image-text) pairs, which, together with this work, will represent a breakthrough in artwork classification and retrieval. Figure \ref{fig:diagram} shows the explained approach.

\begin{figure*}[ht!]
    \centering
    \includegraphics[width=\linewidth]{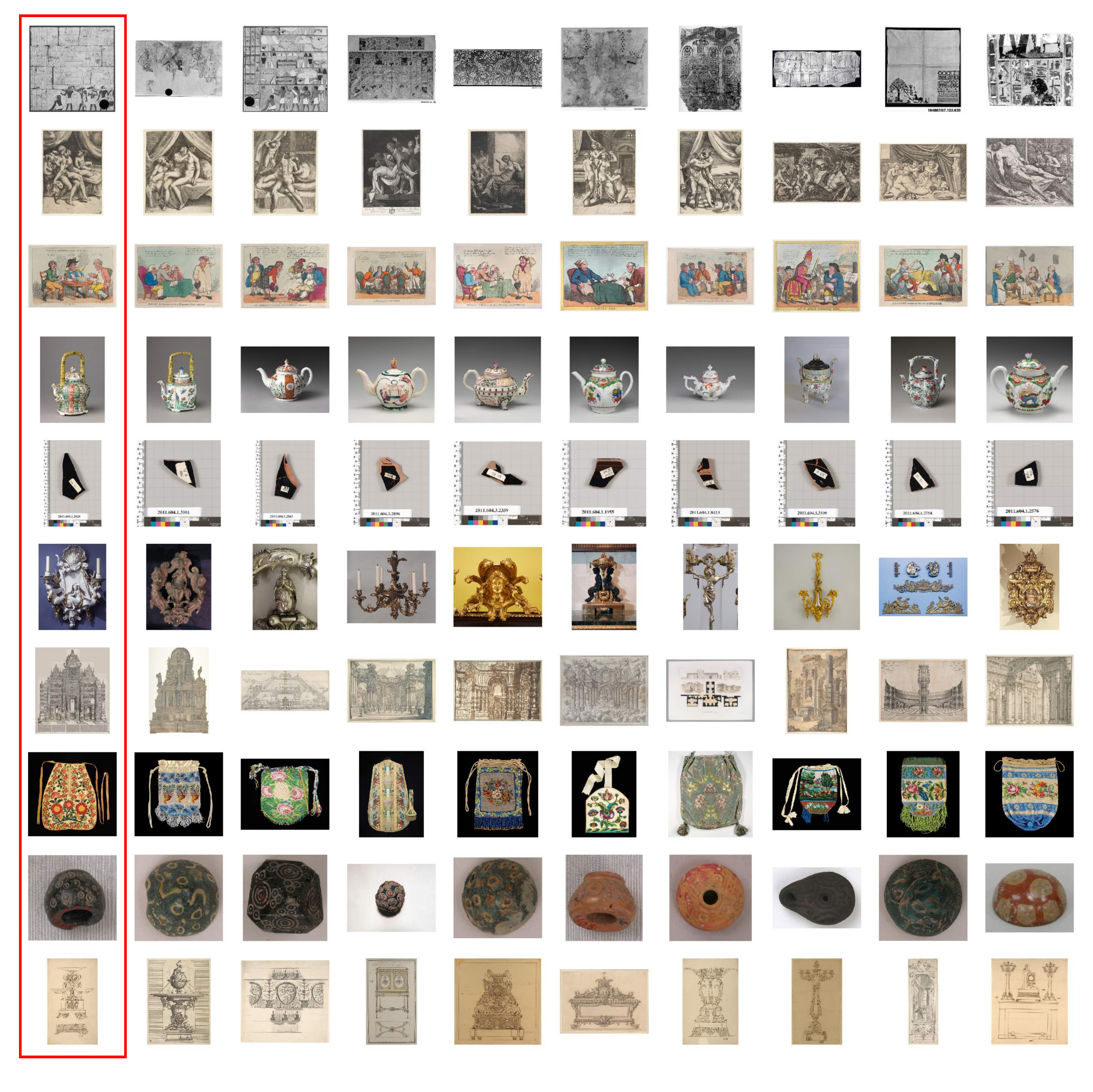}
    \caption{Results for artwork retrieval. We highlighted query images (red bouding box). For each query image we rank 20.000 validation candidates based on cosine similarity, resultant top-9 are shown in each row.
    }
    \label{fig:results-ret-top9}
\end{figure*}

%%%%%%%%%%%%%%%

{\small
\bibliographystyle{ieee_fullname}
\bibliography{egbib.bib}
}

\end{document}